\documentclass[letterpaper]{article}
\usepackage[preprint]{aaai2027}
\usepackage[hyphens]{url}
\usepackage{graphicx}
\usepackage{natbib}
\usepackage{algorithm}
\usepackage{algorithmic}
\usepackage{booktabs}
\usepackage{makecell}
\usepackage{multirow}
\usepackage{mathtools}
\usepackage{amssymb}
\usepackage{xcolor}
\usepackage{float}
\definecolor{draftgray}{RGB}{90,90,90}
\newcommand{\draftfigurebox}[2]{%
  \fbox{\begin{minipage}[c][#1][c]{0.94\linewidth}
    \centering\color{draftgray}\small #2
  \end{minipage}}}

\title{SkyAgent: Revisable Target Memory for Continuous Aerial Search}
\author{
Kaibo Yang\textsuperscript{1,2,3},
Xingyu Wu\textsuperscript{1,2,3},
Enze Zhu\textsuperscript{1,2,3},
Yajie Chai\textsuperscript{4},
Xuyang Li\textsuperscript{1,2,3,4},\\
Jiahui Cui\textsuperscript{1,2,3},
Xiaoxuan Liu\textsuperscript{1,2},
Lei Wang\textsuperscript{1,2},
Zhan Chen\textsuperscript{1,2}\thanks{Corresponding author.}
}
\affiliations{
\textsuperscript{1}Aerospace Information Research Institute, Chinese Academy of Sciences\\
\textsuperscript{2}Key Laboratory of Target Cognition and Application Technology (TCAT)\\
\textsuperscript{3}School of Electronic, Electrical and Communication Engineering,\\
University of Chinese Academy of Sciences\\
\textsuperscript{4}Baidu Inc.
}

\begin{document}
\maketitle

\begin{abstract}
Continuous aerial search builds target hypotheses from sparse,
viewpoint-dependent observations.
When later evidence contradicts a stored hypothesis, however, the
candidate can remain available to the planner and continue triggering
approach or verification actions.
We call this trajectory-level failure \emph{false candidate persistence}
(FCP).
We equip \textsc{SkyAgent} with \emph{Revisable Target Memory} (RTM),
which associates evidence with persistent candidates, discounts
correlated views, and revises each candidate through
\textsc{Tentative}, \textsc{Confirmed}, and \textsc{Rejected} states.
The key interface binds every generated waypoint to its source candidate:
rejection therefore invalidates stale goals, while conservative
reactivation protects temporarily occluded targets. Across four splits from the CityAVOS, UAV-ON, and OpenUAV benchmarks, our \textsc{SkyAgent} equipped with RTM outperforms all competitors, improving SR by 2.90--6.31 points while substantially reducing Behavior-FCP and post-contradiction travel with a low false-rejection rate. Across 35 matched real-world DJI scenarios per method, RTM consistently improves task success and behavioral reliability, further validating its state-to-action correction under physical deployment conditions.
\end{abstract}

\section{Introduction}
\label{sec:intro}
\begin{figure}[!t]
  \centering
  \IfFileExists{figures/fig1.pdf}
    {\includegraphics[width=0.47\textwidth]{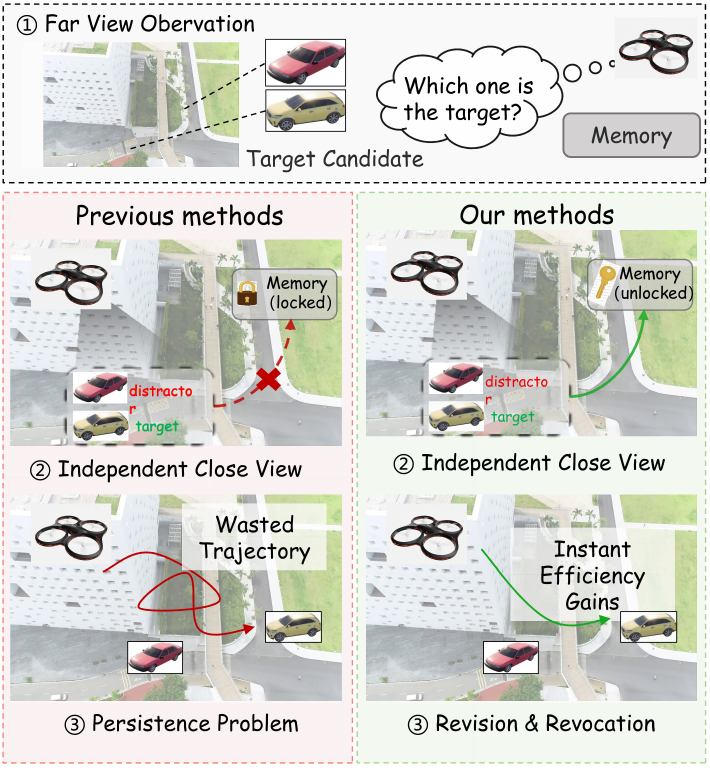}}
    {\IfFileExists{figures/fig1.pdf}
      {\includegraphics[width=0.47\textwidth]{figures/fig1.pdf}}
      {\draftfigurebox{1.45in}{%
        \textbf{Figure 1 placeholder — three columns A/B/C, two rows.}\\[3pt]
        \textit{A}: far positive cue for static distractor $C_1$; both rows
        log $C_1$, $b{=}+0.6$, fly toward it.\\
        \textit{B}: independent side-angle negative view.
        \textbf{Append-only}: $C_1$ unchanged, stale route continues.
        \textbf{RTM}: $b{=}{-}0.4$, state $\to$\textsc{Rejected},
        waypoint canceled.\\
        \textit{C}: true target $C_2$ appears.
        \textbf{Append-only}: drone loops near $C_1$ (red wasted arc),
        episode fails.
        \textbf{RTM}: direct route to $C_2$, $b{=}+0.7$,
        \textsc{Confirmed}, success.}}}
  \caption{\textbf{False candidate persistence in continuous aerial search.}
    A static distractor looks like the target from an early viewpoint
    (\textit{A}) and is contradicted by an independent later view
    (\textit{B}).
    An append-only state keeps $C_1$ active; the drone wastes route on a
    known-wrong candidate (\textit{C}, top).
    RTM rejects $C_1$, cancels its pending waypoint, and resumes search
    toward the true target without erasing the audit trail (\textit{C},
    bottom).}\label{fig:problem}
\end{figure}
In a representative aerial search, a distant rectangular region is logged
as a target candidate.
A later side-angle view reveals that the region is a facade reflection,
not the target.
This revision problem is central to continuous aerial object search: the
physical target is static, but evidence about each candidate shifts
as the vehicle approaches, changes viewpoint, or encounters occlusion
\citep{ji2025cityavos,xiao2025uavon,wang2024openuav}.
The agent rarely sees decisive evidence in a single frame; a distant
shape, a brief gap in foliage, or an intermittent specular return may be
all that a long trajectory provides.

Existing aerial agents have made progress on two fronts.
\emph{Memory-based} methods retain observations across frames through
asynchronous maps, octree representations, or filtered
write-in~\citep{zhang2026apex,zhou2026octmem,ji2025cityavos}.
\emph{Verification-based} methods seek additional viewpoints before
committing to a stop or reconsider a selected snapshot before
executing it~\citep{jang2026contextnav,cao2025cognav,ren2024explore,
wang2026scope}.
Both fronts focus on the \emph{forward} direction: building or validating
confidence before a decision is made.
We focus on what happens \emph{after}: when a later view contradicts
a candidate already sitting in persistent memory, can the system
actually change that memory and cancel the goals it generated?
In the systems we study, the answer is often no---the candidate remains
available to planning even after a closer view proves it wrong.

We define \emph{false candidate persistence} (FCP) as a trajectory-level
failure: after a candidate receives sufficient reference contradiction,
it stays in the active plan (State-FCP) or again induces approach,
revisit, or confirmation actions (Behavior-FCP).
FCP is not a transient detector error, which is a wrong observation in
a single frame; it is a known-wrong hypothesis that continues to consume
distance, time, and planning cycles through stale memory state.
Figure~\ref{fig:problem} illustrates this failure and its correction.

The fix is not more storage.
It is different memory semantics: candidates should be revisable
hypotheses, not committed facts.
A candidate must retain both supporting and opposing evidence, discount
redundant adjacent frames, and expose an explicit lifecycle state to
the planner.
Changing a score alone is not enough; when a candidate is rejected,
every pending waypoint it sourced must be invalidated before the next
planning cycle.
At the same time, rejection must not be permanent, because temporary
occlusion and low-quality frames can produce misleading negative evidence.

We equip the \textsc{SkyAgent} execution backbone with
\emph{Revisable Target Memory} (RTM).
\emph{Candidate Evidence Memory} associates observations with persistent
candidates, groups them by independent viewpoint, and accumulates
signed evidence.
\emph{Revise-and-Suppress} maps accumulated evidence to lifecycle states
through a state-dependent transition function and cancels all stale
waypoints on rejection.
The existing fast search policy, event trigger, inspection skills, and
safety-projected controller remain as execution infrastructure;
candidate truth state is managed exclusively through RTM.

Our contributions are three:
(1)~we identify false candidate persistence (FCP) as a trajectory-level
failure distinct from transient detection errors, and introduce
complementary state-, behavior-, and cost-level measures that separate a
stored false hypothesis from one that still affects flight;
(2)~we propose Revisable Target Memory (RTM), which couples candidate
lifecycle revision to planner binding: rejecting a candidate invalidates
every unexecuted goal it sourced, while conservative reactivation permits
recovery after new independent positive evidence;
(3)~we evaluate RTM through natural FCP prevalence, controlled comparisons
and mechanism ablations on three aerial benchmarks, and 35 matched DJI
scenarios per method (Table~\ref{tab:main}).

\section{Related Work}
\label{sec:related}

\paragraph{Aerial target search and persistent memory.}
Classical aerial target-search systems improve coverage through
visibility-aware viewpoint selection, hierarchical planning, probabilistic
target beliefs, or sparse history maps
\citep{luo2024star,ray2024human,hou2026slider}.
Semantic aerial navigation has since moved from instruction-following
benchmarks such as AerialVLN, OpenUAV, and CityNav
\citep{liu2023aerialvln,wang2024openuav,Lee_2025_ICCV}
to city-scale planners with topo-metric, annotated-map, hierarchical-memory,
and VLM reasoning modules
\citep{gao2024stmr,zhang2025mapnav,zhang2025citynavagent,cai2025flightgpt}.
Training-free aerial frameworks include SPF's image-space waypoint grounding
and PSC-AVDN's parsing--search--confirmation with structured spatial memory;
HTNav and LookasideVLN instead learn tiered or direction-aware policies
\citep{hu2025spf,qi2026pscavdn,fan2026htnav,ning2026lookaside}.
The adjacent aerial object-goal line introduces open-world target
descriptions and long-horizon search in UAV-ON and CityAVOS
\citep{xiao2025uavon,ji2025cityavos}, with APEX, OctMem-Agent, NavFoM, and
RAVEN scaling retrieval through asynchronous maps, octrees, foundation-model
memory, or voxel--ray representations
\citep{zhang2026apex,zhou2026octmem,zhang2026navfom,kim2025raven}.

Persistent memory is also central to ground ObjectNav and lifelong
navigation.
VLMaps and ConceptGraphs attach open-vocabulary semantics to metric maps or
3D scene graphs \citep{huang2023visual,huang2023conceptgraphs};
MemoNav, 3D-Mem, and AstraNav-Memory compress observations into working,
snapshot, or image-centric long-term memories
\citep{li2024memonav,yang2025threedmem,hu2026astranav}.
GOAT-Bench explicitly evaluates memory across a sequence of heterogeneous
goals, while TrajRAG retrieves geometric--semantic experience accumulated
across episodes \citep{khanna2024goat,wang2026trajrag}.
These methods chiefly improve representation, retention, retrieval, or
coverage.
RTM addresses a downstream failure mode: after new evidence contradicts a
stored candidate, whether that candidate remains available to the planner
and what happens to every goal it already generated.

\paragraph{Revision and verification.}
Active perception acquires informative views before committing to a
decision~\citep{bajcsy1988active}; APRR applies this principle to refine the
last-mile viewpoint for precise ObjectNav \citep{qin2025aprr}.
Recent navigation agents verify different objects of uncertainty.
PSC-AVDN confirms a candidate region after a structured search
\citep{qi2026pscavdn}; Context-Nav scores memory-conditioned viewpoints,
CogNav models a cognitive action process, and SCOPE revisits promising
locations before committing
\citep{jang2026contextnav,cao2025cognav,wang2026scope}.
UniGoal uses graph matching for goal-conditioned aerial decisions, while
WMNav corrects short-horizon actions through a learned world model
\citep{yin2025unigoal,nie2025wmnav}.
Explore-Until-Confident calibrates stopping in embodied question answering
rather than target navigation \citep{ren2024explore}.
CARe is especially close: it combines confidence and multi-view consistency
to replan decisions induced by an erroneous pre-explored semantic map
\citep{ko2025care}.
Classical multiple-hypothesis tracking provides a related probabilistic
view of preserving alternatives under ambiguous association
\citep{blackman2004mht}.
RTM does not claim candidate lifecycles or replanning in isolation.
Its distinction is a candidate-local revision contract: every waypoint
records its source candidate, and rejection invalidates all unexecuted
goals sourced from that candidate rather than only reconsidering the next
action.

\paragraph{Online agent memory correction.}
VLM self-correction can revise semantic grounding errors under targeted
prompts~\citep{liao2025selfcorrection}, while Reflexion converts task
feedback into linguistic episodic memory for later trials
\citep{shinn2023reflexion}.
Neither mechanism specifies how a continuously acting navigation system
should retract the downstream commitments of a disproved perceptual
hypothesis.
RTM instead operates online and candidate-locally: it retains supporting and
contradicting observations for audit, changes a candidate's availability to
planning during the same episode, cancels its still-pending waypoints, and
permits recovery only after independent positive views arrive.

\section{Problem Setup}
\label{sec:formulation}

\paragraph{Task.}
Single-frame object detection assumes each observation is independent.
Continuous aerial search instead builds persistent candidate hypotheses
from a stream of partial, viewpoint-dependent observations.
We therefore define the candidate representation to support evidence
accumulation and lifecycle revision.
At each step $t$, the agent receives an observation $o_t$, and the
perception front end produces an evidence item $e_t$:
\begin{equation*}
\begin{aligned}
o_t &= (I_t,\,p_t,\,q),\\
e_t &= (R_t,\,\hat{x}_t,\,f_t,\,y_t,\,\rho_t).
\end{aligned}
\end{equation*}
Here $I_t$ is the current image, $p_t$ is the vehicle pose, and $q$ is
an image or semantic target query.
The candidate region $R_t$ contains a crop or mask, estimated 3-D
location $\hat{x}_t$, multimodal feature $f_t$, signed query-relative
evidence $y_t\in[-1,1]$, and quality $\rho_t\in[0,1]$.
Thus every evidence item refers to one localized hypothesis rather than
assigning an image-level score to multiple candidates.
We study physically static targets throughout.
A candidate's spatial location does not change, but its identity
judgment and associated belief score are updated as new evidence arrives.

\paragraph{Candidate state.}
A binary active/inactive flag is insufficient because approach,
final verification, and complete removal require different planner
permissions.
The $i$-th persistent candidate is therefore represented as
\begin{equation}
m_i^t=\bigl(x_i,\,\mathcal{E}_i^{+,t},\,\mathcal{E}_i^{-,t},\,
b_i^t,\,z_i^t,\,n_i^{+,\mathrm{new}}\bigr),
\label{eq:candidate}
\end{equation}
where $x_i$ is its spatial identity (centroid and uncertainty
ellipsoid), $\mathcal{E}_i^+$ and $\mathcal{E}_i^-$ hold independent
supporting and opposing evidence items, $b_i\in[B_{\min},B_{\max}]$
is a scalar belief score, and
$z_i\in\{\textsc{Tentative},\textsc{Confirmed},\textsc{Rejected}\}$
is its lifecycle state, and $n_i^{+,\mathrm{new}}$ counts independent
positive groups observed since its latest rejection.
\textsc{Confirmed} grants approach and benchmark-defined final
verification; task success is still determined by the benchmark's
stopping rule.

\paragraph{Reference event sets.}
To avoid defining evaluation success with the method's own internal
score, we freeze two external event sets before examining results.
$\mathcal{F}_{\mathrm{ref}}$ contains false candidates with a reference
contradiction time $t_i^\star$, obtained from simulator identity labels
or blinded human annotation.
$\mathcal{R}_{\mathrm{ref}}$ contains true-target recovery episodes with
a degraded or occluded interval followed by new independent positive
views.
The sets are disjoint: temporary loss of a true target is not an FCP
failure type.

\paragraph{FCP measures.}

\noindent\textbf{State-FCP} records whether candidate $i$ stays in the
active plan, while \textbf{Behavior-FCP} records whether it triggers a
concrete action:
\begin{align}
\mathrm{SFCP}_i
  &= \mathbb{I}\!\left[
       \exists t \ge t_i^\star+\Delta:
       i\in\mathcal{C}^{\mathrm{plan}}_t
     \right],
  \label{eq:sfcp}\\
\mathrm{BFCP}_i
  &= \mathbb{I}\!\left[
       \exists t \ge t_i^\star+\Delta:
       a_t\rightsquigarrow i
     \right].
  \label{eq:bfcp}
\end{align}
Here $a_t\rightsquigarrow i$ means that action $a_t$ approaches,
revisits, or confirms candidate $i$.
Over $N_F=|\mathcal{F}_{\mathrm{ref}}|$ eligible events, the reported rates
are
\begin{equation*}
\begin{aligned}
\mathrm{SFCPR}
  &= \frac{1}{N_F}\sum_{i\in\mathcal{F}_{\mathrm{ref}}}\mathrm{SFCP}_i,\\
\mathrm{BFCPR}
  &= \frac{1}{N_F}\sum_{i\in\mathcal{F}_{\mathrm{ref}}}\mathrm{BFCP}_i.
\end{aligned}
\end{equation*}

\noindent\textbf{FCP Waste} (distance and time consumed after
contradiction): total distance and time attributed to actions
$a_t\rightsquigarrow i$ for $t\ge t_i^\star+\Delta$.

Persistence window $\Delta$, source-ID/action-geometry attribution, and
annotation rules are fixed before test evaluation.
We stratify false candidates into same-category distractors, reflections,
repeated texture, and scale/viewpoint confusions; aggregate results use
all eligible events.

\paragraph{Recovery measures.}
For each $j\in\mathcal{R}_{\mathrm{ref}}$, let
\begin{equation*}
\begin{aligned}
\mathrm{FR}_j
 &=\mathbb{I}\!\left[\exists t\in[t_j^-,t_j^+]:
        z_j^t=\textsc{Rejected}\right],\\
\mathrm{OR}_j
 &=\mathbb{I}\!\left[\exists t\in(t_j^+,t_j^{\mathrm{end}}]:
        j\in\mathcal{C}^{\mathrm{plan}}_t\right].
\end{aligned}
\end{equation*}
Here $[t_j^-,t_j^+]$ is the frozen degradation interval and
$t_j^{\mathrm{end}}$ closes the recovery window after new positive views.
We report
$\mathrm{FRR}=N_R^{-1}\sum_j\mathrm{FR}_j$ and
$\mathrm{ORR}=N_R^{-1}\sum_j\mathrm{OR}_j$, where
$N_R=|\mathcal{R}_{\mathrm{ref}}|$.

\begin{figure*}[!t]
  \centering
  \IfFileExists{figures/fig2.pdf}
    {\includegraphics[width=0.999\textwidth]{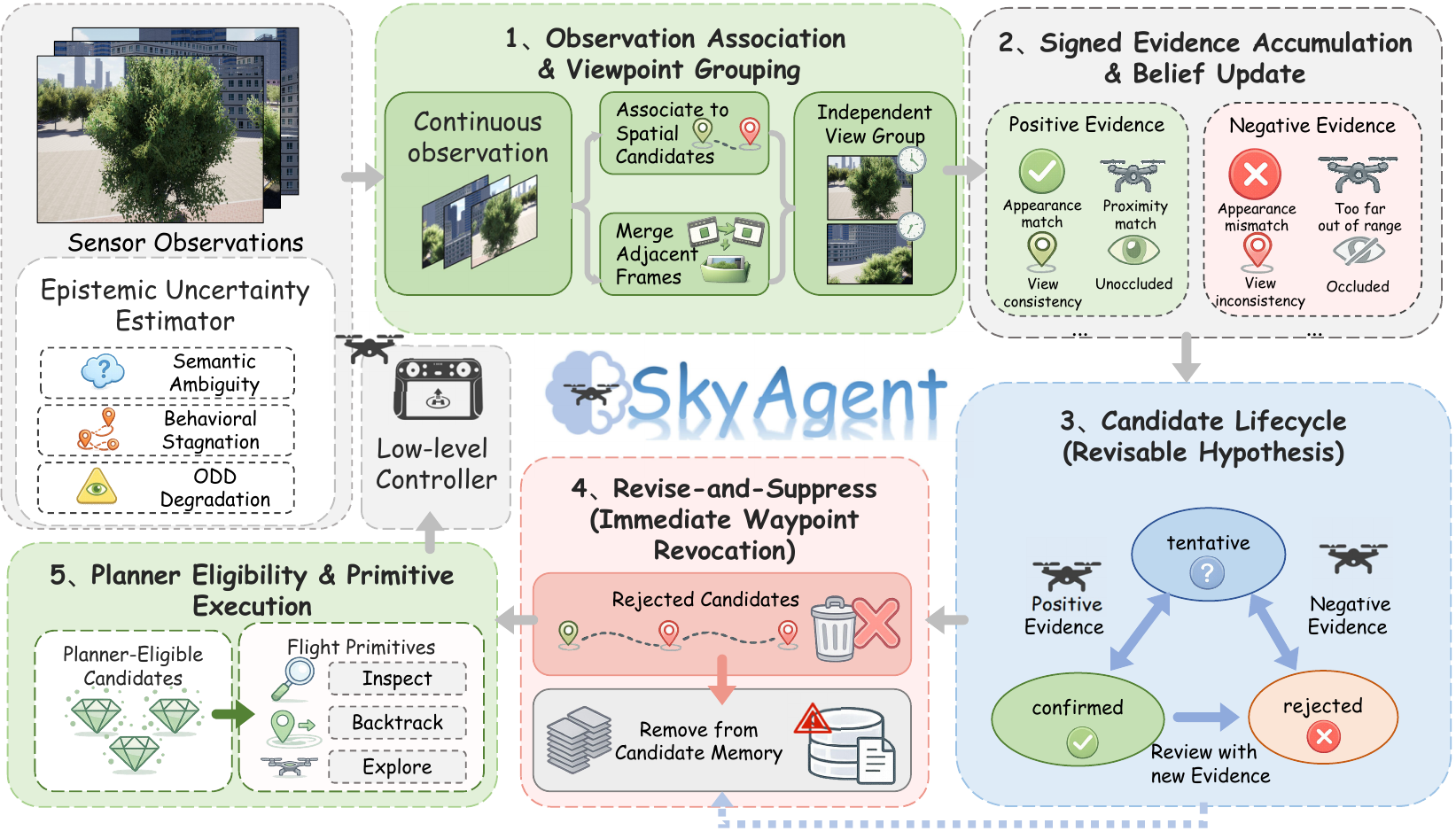}}
    {\IfFileExists{figures/fig2.pdf}
      {\includegraphics[width=0.98\textwidth]{figures/fig2.pdf}}
      {\draftfigurebox{1.35in}{%
        \textbf{Figure 2 placeholder: five-block RTM pipeline.}\\[2pt]
        Block~1: observation filmstrip (t-3 to t) + target query.\\
        Block~2: association into C1/C2; bounded viewpoint-group aggregates
        g1, g2, and g3.
        \textit{Audit log records failure type for Q2 stratification.}\\
        Block~3: signed evidence ledger with +/− chips and score sparklines;
        C1 falls to $b{=}{-}0.3$, C2 rises to $b{=}+0.6$.\\
        Block~4: state machine; C1 transitions Tentative$\to$Rejected,
        C2 Tentative$\to$Confirmed.\\
        Block~5: C1 rejected (red $\times$), waypoint W3 canceled;
        C2 confirmed (green), waypoint W4 active.\\
        Gray execution strip below all blocks: Inspect / Approach / Explore
        $\to$ safety controller.}}}
  \caption{\textbf{Revisable Target Memory.}
    Candidate Evidence Memory groups sequential observations by independent
    viewpoint and accumulates signed evidence for each persistent candidate.
    Revise-and-Suppress updates each candidate's lifecycle through a
    state-dependent transition and propagates rejection to the planner
    by invalidating every pending waypoint sourced from the rejected
    candidate.}\label{fig:method}
\end{figure*}

\section{Revisable Target Memory}
\label{sec:method}

RTM rests on a single interface: every waypoint records the candidate
that sourced it, and rejecting that candidate invalidates all of its
unexecuted goals.
This couples memory revision to planning: the system does not merely
update a score, but propagates rejection through the goal queue.
RTM implements this interface through two modules (Figure~\ref{fig:method}).
\emph{Candidate Evidence Memory} links each observation to a persistent
spatial candidate, groups correlated frames into independent viewpoints,
and accumulates signed evidence.
\emph{Revise-and-Suppress} maps that evidence to a lifecycle state through
a state-dependent transition function and cancels every stale waypoint
the moment a candidate is rejected.
The planner then reads only candidates that are not rejected.

\subsection{Candidate Evidence Memory}

\paragraph{Association.}
Each new observation must be linked to an existing candidate or spawn
a new one.
RTM computes a weighted association score
\begin{equation}
A(e_t,m_i^t)
  = w_x A_x + w_f A_f + w_\delta A_\delta,
\label{eq:association}
\end{equation}
with components
\begin{equation*}
\begin{aligned}
A_x
  &= \exp\!\left(
       -\frac{\lVert\hat{x}_t-x_i\rVert_2^2}{\sigma_x^2}
     \right),\\
A_f
  &= \cos\!\left(f_t,\bar{f}_i^t\right),\\
A_\delta
  &= \exp\!\left(-\frac{\Delta t}{\tau_\delta}\right).
\end{aligned}
\end{equation*}
The three terms measure spatial proximity, appearance similarity, and
temporal recency, respectively.
Here $\bar{f}_i^t$ is the running mean feature of candidate $i$, and
$\Delta t$ is the number of steps since $i$ was last updated.
The best-scoring candidate is selected: $i^\star = \arg\max_i
A(e_t,m_i^t)$.
If $A(e_t,m_{i^\star}^t) < \tau_{\mathrm{assoc}}$, a new candidate is
created with $b=0$ and $z=\textsc{Tentative}$.
The maximization includes rejected records, which remain in the
association and audit store even though they are unavailable to planning.
Before creating a new ID, RTM also checks rejected records with the
calibrated spatial and appearance gates but without temporal recency;
elapsed time alone therefore cannot bypass the reactivation guard.
Every create, association, and unassigned event is logged for audit.
Failure-source labels are assigned only in the frozen offline evaluation;
they are not inputs to association, evidence accumulation, or state
transition.
An oracle-association ablation (Table~\ref{tab:ablation}) separately
estimates how much of the remaining error comes from candidate tracking
rather than from the revision mechanism itself.

\paragraph{Independent viewpoint groups.}
Hovering near a false object long enough to collect many similar frames
should not raise its confidence score.
Two observations belong to the same viewpoint group when all four
conditions hold:
\begin{equation}
\begin{alignedat}{2}
\Delta\theta_{it} &\le \theta_{\max},
&\qquad \Delta d_{it} &\le d_{\max},\\
\Delta h_{it} &\le h_{\max},
& \Delta t &\le T_{\max}.
\end{alignedat}
\label{eq:viewgroup}
\end{equation}
where $\Delta\theta_{it}$ is the bearing change from $p_t$ toward $x_i$
relative to the group's representative view, $\Delta d_{it}$ is range
change, and $\Delta h_{it}$ is altitude change.
All observations in a group maintain one bounded aggregate contribution;
new correlated frames update that aggregate rather than casting additional
votes.
The counts $n_i^+$ and $n_i^-$ track the number of distinct groups in
$\mathcal{E}_i^+$ and $\mathcal{E}_i^-$ separately; they gate
reactivation in Equation~\ref{eq:lifecycle}.

\paragraph{Signed evidence.}
A frozen query-conditioned match model $h_\phi$, calibrated on a
held-out split, produces a probability that is converted to the signed
evidence $y_t$ defined in the evidence item:
\begin{equation*}
\begin{aligned}
r_t &= h_\phi(R_t,q),
& y_t &= 2r_t-1.
\end{aligned}
\end{equation*}
Here $r_t\in[0,1]$.
Positive values support the candidate's target identity; negative values
contradict it; values within $(-\epsilon, \epsilon)$ are discarded as
uninformative.
The magnitude $|y_t|$ already encodes semantic confidence.
For viewpoint group $g$, define the quality-weighted bounded contribution
$s_{ig}^t$ and update the belief by replacing the previous contribution
of that group:
\begin{equation}
\begin{aligned}
s_{ig}^t
  &=\frac{\sum_{k\in g^t}\rho_k y_k}
          {\max(1,\sum_{k\in g^t}\rho_k)},\\
b_i^{t+1}
  &=\operatorname{clip}_{[B_{\min},B_{\max}]}
    \!\left(b_i^t+s_{ig}^{t+1}-s_{ig}^{t}\right).
\end{aligned}
\label{eq:update}
\end{equation}
For a newly created group, $s_{ig}^{t}=0$ before its first update.
Because $|s_{ig}^t|\le1$, collecting arbitrarily many correlated frames
cannot increase a group's total evidence mass.
Equation~\ref{eq:update} is an interpretable evidence accumulator;
we do not claim a calibrated Bayesian posterior.

\subsection{Revise-and-Suppress}

\paragraph{State-dependent lifecycle.}
Transitions depend on the current score \emph{and} the current state,
not on the score alone.
Define the four transition guards
\begin{equation*}
\begin{aligned}
\gamma_i^{\mathrm{conf}}
  &\equiv
    (z_i^t\ne\textsc{Rejected})
    \land (b_i^{t+1}\ge\tau_{\mathrm{pos}}),\\
\gamma_i^{\mathrm{rej}}
  &\equiv
    (z_i^t\ne\textsc{Rejected})
    \land (b_i^{t+1}\le\tau_{\mathrm{neg}}),\\
\gamma_i^{\mathrm{dem}}
  &\equiv
    (z_i^t=\textsc{Confirmed})
    \land
    (b_i^{t+1}<\tau_{\mathrm{pos}}-\delta_{\mathrm{hyst}}),\\
\gamma_i^{\mathrm{react}}
  &\equiv
    (z_i^t=\textsc{Rejected})
    \land (b_i^{t+1}\ge\tau_{\mathrm{react}})\\
  &\quad{}\land
    (n_i^{+,\mathrm{new}}\ge n_{\mathrm{react}}).
\end{aligned}
\end{equation*}
Applied in the order shown below, the lifecycle update is
\begin{equation}
z_i^{t+1} =
\begin{cases}
\textsc{Confirmed},
  & \gamma_i^{\mathrm{conf}},\\
\textsc{Rejected},
  & \gamma_i^{\mathrm{rej}},\\
\textsc{Tentative},
  & \gamma_i^{\mathrm{dem}}\lor\gamma_i^{\mathrm{react}},\\
z_i^t,
  & \text{otherwise}.
\end{cases}
\label{eq:lifecycle}
\end{equation}
Here $\tau_{\mathrm{neg}}<\tau_{\mathrm{pos}}$ forms the hysteresis band;
$\delta_{\mathrm{hyst}}$ prevents \textsc{Confirmed} from flickering
under small score drops; $\tau_{\mathrm{react}}>\tau_{\mathrm{neg}}$
sets a stricter recovery bar; and $n_{\mathrm{react}}$ requires at least
that many new independent positive viewpoint groups accumulated
\emph{after} the most recent rejection.
A rejected candidate therefore requires genuinely new independent
positive evidence to recover, not a minor score oscillation.
On entry to \textsc{Rejected}, RTM resets
$n_i^{+,\mathrm{new}}$ to zero.
Subsequent independent positive groups increment it once each.
Reactivation deliberately returns the candidate only to
\textsc{Tentative}; confirmation requires a later update, preventing a
single transition from restoring planning authority and confirming the
target simultaneously.

\textsc{Tentative} candidates may request additional viewpoints but
cannot terminate the episode.
\textsc{Confirmed} candidates may be approached and checked under the
benchmark's stopping rule.
\textsc{Rejected} candidates remain in the audit log but leave the
active candidate set; they may return to \textsc{Tentative} only via
the reactivation guard in Equation~\ref{eq:lifecycle}.

\paragraph{Planner binding.}
The planner reads only non-rejected candidates:
\begin{equation}
\mathcal{C}^{\mathrm{plan}}_t =
\{m_i^t \mid z_i^t \ne \textsc{Rejected}\}.
\label{eq:eligible}
\end{equation}
Every waypoint stores its source candidate ID at generation time.
When candidate $i$ enters \textsc{Rejected}, every unexecuted waypoint
tagged with ID $i$ is canceled before the next planning cycle begins.
An action already under safety-critical execution is completed or
terminated by the safety controller rather than interrupted by memory.
If eligible candidates remain, planning continues from the remaining
set; otherwise the agent resumes coverage search.
This cancellation is the step that verify-before-act methods do not
take: they replan the next action, but leave previously queued
candidate-bound goals intact.

\begin{table*}[t]
\centering
\footnotesize
\setlength{\tabcolsep}{4.2pt}
\renewcommand{\arraystretch}{0.96}
\begin{tabular}{@{}llcccccc@{}}
\toprule
\textbf{Benchmark} & \textbf{Method}& \textbf{SR}$\uparrow$
  & \textbf{Eff.}$\uparrow$
  & \textbf{Dist.}$\downarrow$
  & \textbf{BFCPR}$\downarrow$
  & \textbf{Waste}$\downarrow$
  & \textbf{FRR}$\downarrow$\\
\midrule
\multirow{6}{*}{\makecell[l]{CityAVOS-Hard\\(SkyAgent $n{=}380$)}}
& RE~\citep{ji2025cityavos}
  & 5.05 & 2.47 & 674.16 & \textnormal{N/A} & \textnormal{N/A} & \textnormal{N/A}\\
& FBE~\citep{ji2025cityavos}
  & 8.07 & 6.34 & 628.70 & \textnormal{N/A} & \textnormal{N/A} & \textnormal{N/A}\\
& STMR~\citep{ji2025cityavos}
  & 20.96 & 13.23 & 547.77 & \textnormal{N/A} & \textnormal{N/A} & \textnormal{N/A}\\
& PRPSearcher (Hard subset)~\citep{ji2025cityavos}
  & 29.62 & 16.65 & 507.60 & \textnormal{N/A} & \textnormal{N/A} & \textnormal{N/A}\\
& Append-only SkyAgent
  & 33.95 & 23.16 & 431.13 & 45.18 & 120.76 & \textnormal{N/A}\\
& \textbf{SkyAgent + RTM (Hard)}
  & \textbf{40.26} & \textbf{30.14} & \textbf{349.14}
  & \textbf{15.27} & \textbf{44.21} & \textbf{2.91}\\
\midrule
\multirow{3}{*}{\makecell[l]{CityAVOS-Medium\\(SkyAgent $n{=}380$)}}
& PRPSearcher (Medium subset)~\citep{ji2025cityavos}
  & 42.89 & 36.68 & 362.11 & \textnormal{N/A} & \textnormal{N/A} & \textnormal{N/A}\\
& Append-only SkyAgent (Medium)
  & 45.26 & 36.85 & 360.48 & 20.39 & 55.23 & \textnormal{N/A}\\
& \textbf{SkyAgent + RTM (Medium)}
  & \textbf{48.16} & \textbf{40.11} & \textbf{329.13}
  & \textbf{7.69} & \textbf{24.82} & \textbf{1.94}\\
\midrule
\multirow{6}{*}{\makecell[l]{UAV-ON\\($n{=}1{,}000$)}}
& Random~\citep{zhou2026octmem}
  & 3.70 & 2.66 & 596.53 & \textnormal{N/A} & \textnormal{N/A} & \textnormal{N/A}\\
& CLIP-H~\citep{zhou2026octmem}
  & 6.20 & 4.15 & 547.32 & \textnormal{N/A} & \textnormal{N/A} & \textnormal{N/A}\\
& AOA-F~\citep{zhou2026octmem}
  & 7.30 & 4.06 & 550.29 & \textnormal{N/A} & \textnormal{N/A} & \textnormal{N/A}\\
& OctMem-Agent~\citep{zhou2026octmem}
  & 19.50 & 6.37 & 474.00 & \textnormal{N/A} & \textnormal{N/A} & \textnormal{N/A}\\
& Append-only SkyAgent
  & 20.30 & 6.53 & 468.72 & 29.95 & 90.97 & \textnormal{N/A}\\
& \textbf{SkyAgent + RTM}
  & \textbf{23.20} & \textbf{8.07} & \textbf{417.86}
  & \textbf{12.32} & \textbf{40.71} & \textbf{3.09}\\
\midrule
\multirow{6}{*}{\makecell[l]{OpenUAV-Hard\\($n{=}816$)}}
& Random Action~\citep{zhang2026navfom}
  & 0.00 & 0.00 & 800.66 & \textnormal{N/A} & \textnormal{N/A} & \textnormal{N/A}\\
& Fixed Action~\citep{zhang2026navfom}
  & 1.10 & 0.71 & 776.22 & \textnormal{N/A} & \textnormal{N/A} & \textnormal{N/A}\\
& CMA~\citep{zhang2026navfom}
  & 3.43 & 3.37 & 684.63 & \textnormal{N/A} & \textnormal{N/A} & \textnormal{N/A}\\
& TravelUAV~\citep{zhang2026navfom}
  & 12.74 & 11.76 & 395.77 & \textnormal{N/A} & \textnormal{N/A} & \textnormal{N/A}\\
& Append-only SkyAgent
  & 18.01 & 7.03 & 558.62 & 35.32 & 120.90 & \textnormal{N/A}\\
& \textbf{SkyAgent + RTM}
  & \textbf{22.06} & \textbf{9.03} & \textbf{489.76}
  & \textbf{14.07} & \textbf{50.41} & \textbf{3.52}\\
\bottomrule
\end{tabular}
\caption{\textbf{Performance on aerial benchmarks.}
SkyAgent rows are paired internal comparisons. External SR and Eff.\ values
are reported by the cited papers; Dist.\ values are outputs of our benchmark
reproduction rather than values copied from their tables. External methods
may use different perception front ends.
SR, Eff., BFCPR, and FRR are percentages; Dist.\ and Waste are meters.}
\label{tab:main}
\end{table*}

\subsection{Integration with Flight Execution}
The fast search policy, event trigger, inspection skills, and
safety-projected controller are retained as execution infrastructure.
They collect observations and execute bounded actions, but candidate
state is read and written exclusively through RTM.
The language router may emit structured evidence judgments or select
\texttt{Inspect}; it cannot directly set $z_i$, reactivate a candidate,
or bypass waypoint cancellation.
The architecture therefore satisfies a separation of concerns: the
revision mechanism can be tested and ablated independently of the
specific router or low-level controller.

\section{Experiments}
\label{sec:experiments}

We evaluate natural FCP prevalence, controlled benchmark effects,
mechanism and recovery, and paired physical transfer.

\begin{figure}[!t]
  \centering
  \IfFileExists{figures/fig3.pdf}
    {\includegraphics[width=\columnwidth]{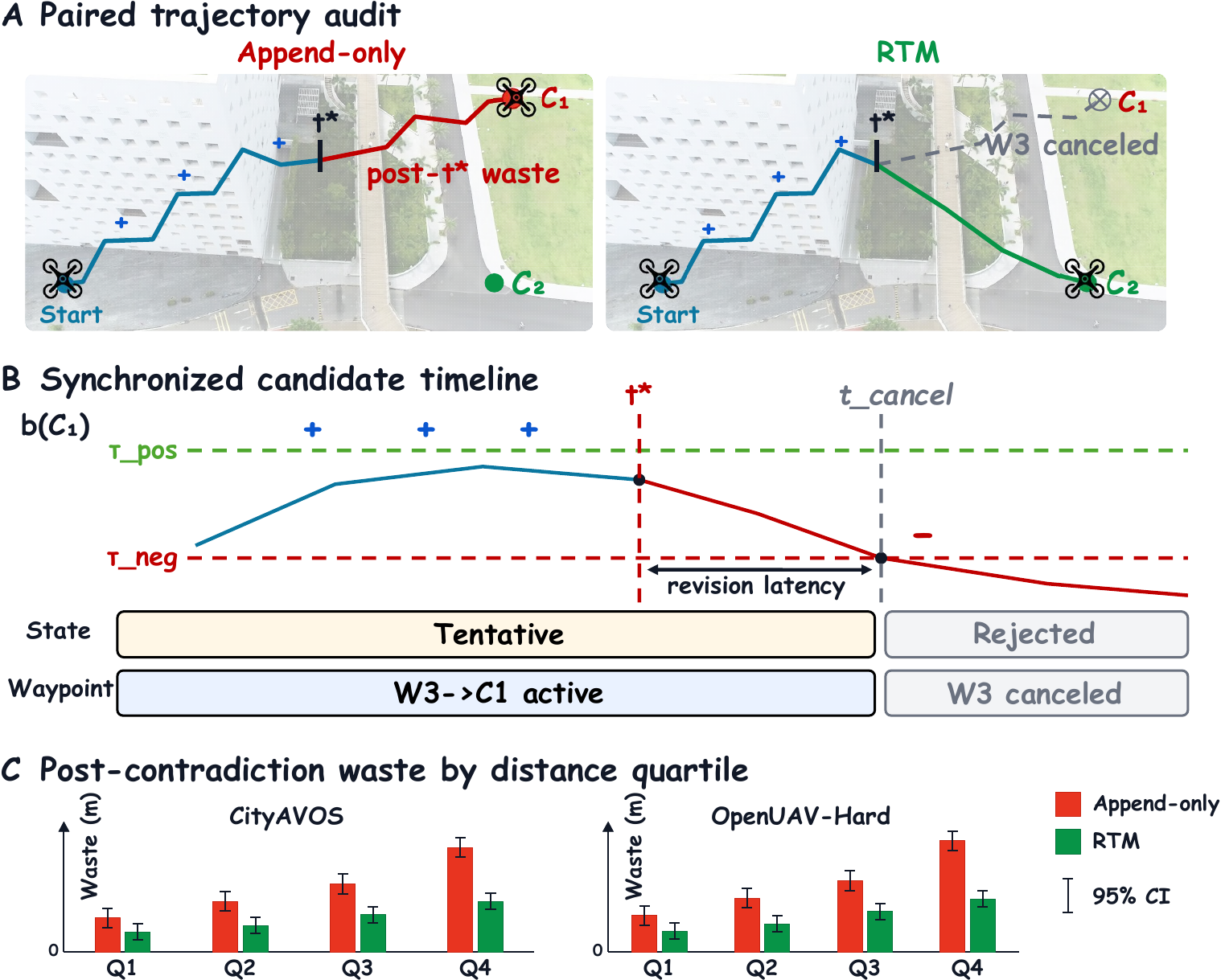}}
    {\IfFileExists{figures/fig3.pdf}
      {\includegraphics[width=\columnwidth]{figures/fig3.pdf}}
      {\draftfigurebox{2.15in}{%
        \textbf{Three-panel mechanism evidence.}\\[3pt]
        \textit{Panel A}: paired top-down trajectories after $t^\star$.\\
        \textit{Panel B}: synchronized candidate score, lifecycle state,
        and waypoint cancellation.\\
        \textit{Panel C}: post-contradiction waste by distance quartile.}}}
  \caption{\textbf{Mechanism audit layout.}
    \textit{Panel A} contrasts append-only and RTM trajectories after the
    reference contradiction at $t^\star$.
    \textit{Panel B} aligns signed evidence, candidate state, and stale-goal
    cancellation.
    \textit{Panel C} defines the distance-quartile summary used for the
    supplementary stress analysis.}
  \label{fig:benchmark_mechanism}
\end{figure}

\subsection{Setup}

\paragraph{Environments.}
We use three public aerial benchmarks.
\textbf{CityAVOS} is used to form 380 Hard tasks and a 380-task Medium
control, matched one-to-one by nearest start-to-goal
distance~\citep{ji2025cityavos}.
\textbf{UAV-ON} contributes 1,000 test episodes from 14 Unreal Engine
environments~\citep{xiao2025uavon}.
\textbf{OpenUAV-Hard} combines 420 Test-Unseen-Object and 396
Test-Unseen-Map trajectories of at least 250\,m under its native
protocol~\citep{wang2024openuav}.
RTM uses only benchmark-standard observations and goal information.

\paragraph{Protocol.}
Within each benchmark, RTM variants share task IDs, targets, perception,
action budgets, controllers, and evaluators; thresholds, prompts, association
rules, and failure handling are fixed on disjoint calibration episodes.
Complete-system baselines use official implementations when available.
SPF and PSC-AVDN are adapted to the common DJI interface without task-specific
training; their table values are measured under our protocol.
Success uses an exact paired binary test; continuous and event measures use
paired, episode-clustered bootstrap intervals.

\paragraph{Metrics.}
We report benchmark-native SR, path efficiency, distance, and time.
Mechanism metrics are SFCPR, BFCPR, post-contradiction Waste, and revision
latency. FRR and ORR measure suppression cost and recovery; association
accuracy and model calls are supporting diagnostics.

\begin{table*}[t]
\centering
\small
\setlength{\tabcolsep}{4.8pt}
\begin{tabular}{@{}lccccccc@{}}
\toprule
\textbf{Variant}
  & \textbf{SR}$\uparrow$
  & \textbf{SFCPR}$\downarrow$
  & \textbf{BFCPR}$\downarrow$
  & \textbf{Waste}$\downarrow$
  & \textbf{FRR}$\downarrow$
  & \textbf{ORR}$\uparrow$
  & \textbf{Latency}$\downarrow$\\
\midrule
Append-only
  & 33.95 & 55.33 & 45.18 & 120.76 & -- & 77.08 & --\\
Verify current action
  & 36.05 & 50.25 & 38.19 & 96.17 & 1.96 & 80.21 & 6.53\\
RTM w/o negative evidence
  & 35.53 & 51.24 & 39.80 & 103.18 & 0.99 & 82.29 & 7.29\\
RTM w/o viewpoint grouping
  & 37.63 & 27.23 & 21.29 & 60.92 & 8.65 & 77.08 & 2.37\\
Revision state only
  & 38.42 & 19.90 & 28.86 & 74.24 & 3.88 & 89.58 & 8.72\\
Permanent rejection
  & 38.68 & 15.42 & 12.44 & 38.26 & 11.65 & 65.63 & 2.52\\
\textbf{Full RTM}
  & \textbf{40.26} & \textbf{16.75} & \textbf{15.27}
  & \textbf{44.21} & \textbf{2.91} & \textbf{92.71} & \textbf{3.19}\\
Full + oracle association
  & 42.11 & 9.80 & 8.82 & 29.50 & 1.92 & 95.83 & 2.54\\
\bottomrule
\end{tabular}
\caption{\textbf{Mechanism ablations on CityAVOS-Hard ($n{=}380$).}
Closed-loop rates use row-specific externally annotated events; the
shared-stream analysis and all event counts are reported separately.
Waste is meters and Latency is steps.}
\label{tab:ablation}
\end{table*}

\begin{figure*}[!t]
\centering
\includegraphics[width=\textwidth]{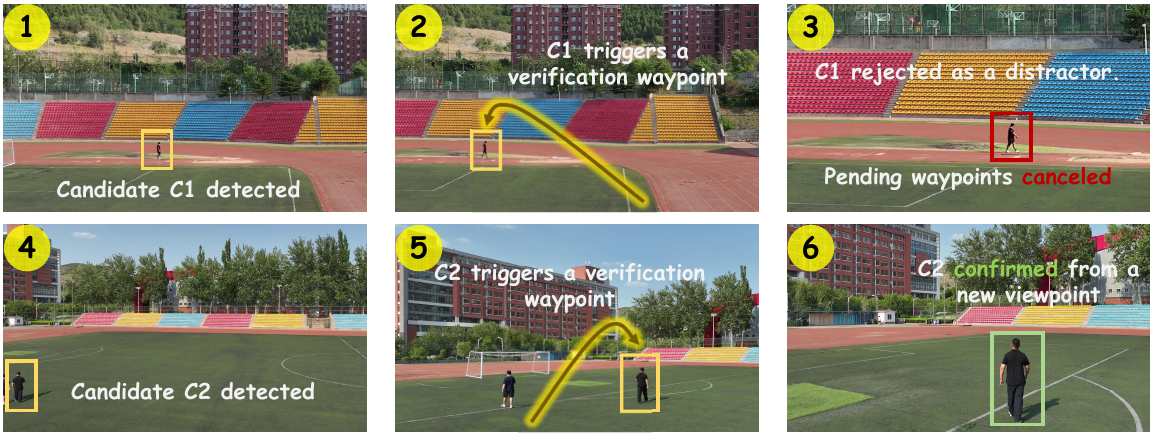}
\caption{\textbf{Representative real-flight revision sequence.}
  Inspection rejects $C_1$ and cancels its queued waypoint; search resumes
  toward $C_2$, confirmed from an independent view. Table~\ref{tab:real}
  reports all paired trials.}
\label{fig:real}
\end{figure*}

\subsection{FCP in Unmodified Search}

Among 380 append-only CityAVOS-Hard episodes, 197 contain a reference event;
89 contradicted candidates remain active (BFCPR 45.18\%), causing 120.76\,m
of post-contradiction travel. Two annotators label creation, the first
reference contradiction, later eligibility, false confirmation, and
attributable distance under fixed $t_i^\star$ rules, with adjudication across
five distractor, appearance, viewpoint, occlusion, and degradation strata.

\subsection{Controlled Benchmark Results}

\paragraph{FCP suppression.}
Across four splits, RTM reduces CityAVOS-Hard BFCPR from
45.18\% to 15.27\% (29.91 points; 66.2\% relative) and reduces
post-contradiction waste from 120.76\,m to 44.21\,m.
The same pattern holds on CityAVOS-Medium (12.70 points and 30.41\,m),
UAV-ON (17.63 points and 50.26\,m), and OpenUAV-Hard (21.25 points and
70.49\,m), while FRR remains 1.94--3.52\%.

\paragraph{Task outcomes.}
RTM improves SR and path efficiency on every split; SR gains are
2.90--6.31 points and peak on ambiguity-rich CityAVOS-Hard.
SkyAgent comparisons use identical paired episodes and complete splits.
CityAVOS baseline values come from the difficulty-stratified 605-task
preprint evaluation \citep{ji2025cityavos}; unavailable stratum counts
are therefore not assigned to those rows.
OpenUAV-Hard baselines combine the published Test-Unseen-Object
($n{=}420$) and Test-Unseen-Map ($n{=}396$) Hard aggregates.
Eff.\ is paper-reported SPL and \textnormal{N/A} marks unavailable
RTM diagnostics.
BFCPR and Waste reductions support revision as the mechanism, although
route changes also shift downstream observations.

\subsection{Mechanism, Recovery, and Stress}

\paragraph{Component isolation.}
Rechecking only the current action leaves BFCPR at 38.19\% and Waste at
96.17\,m, versus 15.27\% and 44.21\,m for Full RTM.
State revision without queued-goal cancellation reaches 28.86\% BFCPR and
74.24\,m Waste, isolating the value of planner binding.
Removing negative evidence approaches append-only behavior; removing
viewpoint grouping raises FRR from 2.91\% to 8.65\%.

Permanent rejection lowers BFCPR but raises FRR to 11.65\% and reduces
ORR to 65.63\%; Full RTM preserves 92.71\% ORR with 2.91\% FRR.
Oracle association reaches 42.11\% SR and 8.82\% BFCPR, identifying
cross-view association as the main remaining bottleneck.

\paragraph{Long-horizon and recovery controls.}
On 816 OpenUAV-Hard trajectories, RTM lowers BFCPR from 35.32\% to 14.07\%
and Waste from 120.90\,m to 50.41\,m while SR rises from 18.01\% to 22.06\%.
The CityAVOS-Hard control confirms the FRR/ORR trade-off;
Figure~\ref{fig:benchmark_mechanism} audits trajectories, candidate states,
and cancellation, with shared-stream and stratified analyses in the supplement.

\subsection{Real-Flight Evaluation}

\begin{table}[!h]
\centering
\scriptsize
\setlength{\tabcolsep}{2.2pt}
\begin{tabular}{@{}lcccccc@{}}
\toprule
\textbf{Method}
  & \textbf{SR}$\uparrow$
  & \textbf{Dist.}$\downarrow$
  & \textbf{BFCPR}$\downarrow$
  & \textbf{FRR}$\downarrow$
  & \textbf{Rev.Lat.(s)}$\downarrow$
  & \textbf{Int.}$\downarrow$\\
\midrule
SPF (TF)
  & 54.29 & 275.68 & 50.00 & 12.50 & 6.31 & 17.14\\
PSC-AVDN (TF)
  & 60.00 & 257.94 & 40.91 & 16.67 & 5.24 & 11.43\\
\midrule
Append-only SkyAgent
  & 57.14 & 264.37 & 45.45 & -- & -- & 14.29\\
\textbf{SkyAgent + RTM}
  & \textbf{71.43} & \textbf{223.81} & \textbf{13.64}
  & \textbf{4.17} & \textbf{2.78} & \textbf{5.71}\\
\bottomrule
\end{tabular}
\caption{\textbf{Matched DJI flights.}
Results use 35 matched scenarios/method (140 trials).
Rates are percentages, Dist.\ is meters, and Rev.Lat.\ is seconds;
TF denotes no task-specific training or fine-tuning.}
\label{tab:real}
\end{table}

With the common backend, training-free SPF and PSC-AVDN reach 54.29/60.00\%
SR and 50.00/40.91\% BFCPR~\citep{hu2025spf,qi2026pscavdn}.
Against append-only, RTM raises SR from 57.14\% to 71.43\%, cuts distance
from 264.37\,m to 223.81\,m, BFCPR from 45.45\% to 13.64\%, and
interventions from 14.29\% to 5.71\%; FRR is 4.17\% and latency 2.78\,s.
The four methods share 35 starts, regions, layouts, weather strata, and
budgets; order is randomized and all outcomes retained across five conditions.
Figure~\ref{fig:real} shows one sequence; Table~\ref{tab:real} covers 140 trials.

\section{Limitations and Conclusion}
\label{sec:conclusion}

RTM assumes reliable cross-view association; merging or splitting candidates
distorts FCP counts, and oracle association estimates this tracking
contribution. The frozen evidence model may retain systematic biases.
Because RTM also changes later routes and observations, closed-loop results
measure system behavior; common-trajectory replay is needed to isolate
revision. CityAVOS-calibrated thresholds may require recalibration for new
cameras or altitudes. The present scope covers static objects, and the tested
flights do not establish a general safety guarantee.

Candidate Evidence Memory aggregates signed evidence across viewpoints, while
Revise-and-Suppress cancels waypoints sourced by rejected candidates.
RTM improves task performance across four paired benchmark splits while
reducing FCP and keeping false rejection low.

\bibliography{references,references_rtm_additions}

\end{document}